\documentclass[11pt]{article}

\usepackage[margin=1in]{geometry}
\usepackage{amsmath,amssymb,booktabs,graphicx}
\usepackage{natbib}
\usepackage{microtype}
\usepackage[colorlinks=true, allcolors=blue]{hyperref}
\usepackage{enumitem}
\graphicspath{{figures/}{manuscript/figures/}}

\title{Transformer-Based Reinforcement Learning for Phase I Dose Allocation}
\author{Shijie Yuan\\
The University of Texas at Austin\\
\texttt{sj.yuan@austin.utexas.edu}}
\date{}

\begin{document}
\maketitle

\begin{abstract}
Phase I dose-finding trials must learn from sparse toxicity observations while limiting exposure to overly
toxic doses. We develop a single reinforcement-learning policy for trials with different numbers of ordered
dose levels. Each dose is represented as a token containing its normalized position, accumulated patient
count, observed dose-limiting toxicities (DLTs), and current-dose indicator. A Transformer encoder summarizes
the dose tokens, trial-conditioned cross-attention produces a fixed-dimensional trial representation, and a
three-action policy chooses de-escalation, staying, or escalation. The final maximum tolerated dose (MTD)
decision is deliberately kept outside the learned policy: after the planned sample size, weighted isotonic
regression and a prespecified safety cap select the dose closest to the target toxicity probability. In the
simulations, toxicity-penalized policies reduce DLT burden while unpenalized policies improve selection
accuracy at the cost of greater toxicity. These results demonstrate a controllable safety--accuracy trade-off
but do not establish superiority over conventional dose-finding designs.
\end{abstract}

\section{Introduction}

Phase I oncology trials must learn an acceptable dose from limited patient cohorts while protecting
participants from excessive toxicity. The operational decision is sequential: after each cohort, the trial
must decide whether to escalate, remain at the current dose, or de-escalate. The statistical target is
usually expressed through a dose-limiting toxicity (DLT) probability $\phi$, but the evidence available at
each decision is sparse and unevenly distributed across dose levels.

For example, consider a trial with six ordered dose levels, $d_1<\cdots<d_6$, and a target DLT probability of
$\phi=0.25$. The trial may begin at $d_1$ and observe no DLTs in the first cohort, supporting escalation to
$d_2$. If subsequent cohorts at $d_2$ and $d_3$ continue to show little toxicity, the allocation rule may
move upward to gather information near the target. Conversely, if two DLTs are observed among three patients
at $d_3$, the next decision should become more conservative: the trial may remain at $d_3$, de-escalate to
$d_2$, or stop if the evidence indicates that even the lowest dose is unsafe. The same observed DLT count can
imply different actions depending on the current dose, the number of patients already treated, the remaining
sample-size budget, and the information collected at neighboring doses. This context dependence is the
central sequential-allocation problem addressed here.

Conventional Phase I designs provide important safety and interpretability properties. Algorithmic and
model-assisted approaches such as BOIN, i3+3, and mTPI-2 use prespecified decision boundaries or
interval-based rules \citep{liu2015boin,liu2020i3,guo2017mtpi2}. Model-based approaches such as CRM impose a
dose--toxicity model \citep{oquigley1990crm,cheung2011dose}, while isotonic approaches provide an alternative
way to enforce monotonicity when estimating a dose close to the target \citep{ivanova2009comparison}. These
conventional dose-finding designs are strong reference methods, but their allocation rules are generally
specified for a fixed dose grid and do not learn a shared representation across trials with different numbers
of dose levels.

Deep reinforcement learning has recently been explored for Phase I dose escalation
\citep{matsuura2023optimal}, creating an opportunity to optimize sequential allocation directly against
trial-level objectives. In the reported formulation, the escalation rule is learned through simulated trials,
as is typical for model development in this setting. Consequently, its operating characteristics are tied to
the prespecified state, action, reward, and scenario-generating distributions, and the extent to which the
learned rule transfers to alternative dose grids or clinical settings remains an empirical question. The
multidimensional policy may also require additional tools for interpretation, and coupling allocation with
final MTD recommendation can make it less direct to separate sequential-allocation error from
final-dose-estimation error.

These considerations motivate a framework with a shared representation across variable numbers of dose levels
and an explicitly prespecified final selection rule. The proposed approach retains the flexibility of learned
sequential allocation while keeping the final clinical recommendation transparent and separately evaluable.

The challenge is to obtain the flexibility of a learned policy without sacrificing transparency or allowing a
training reward to define the final clinical recommendation implicitly. We therefore separate allocation from
final selection. During the trial, a policy learns only the next-cohort action. At planned completion, a
prespecified weighted isotonic-regression rule selects the MTD from observed data. This separation also makes
safety--accuracy trade-offs visible: a policy can be evaluated by DLT burden, incorrect early stopping, and
MTD-selection accuracy rather than by a single reward alone.

We present a Transformer-based reinforcement-learning framework in which dose levels are represented as a
variable-length token sequence and trial-level context supplies a query for cross-attention. The resulting
representation supports a fixed three-action policy for trials with different dose counts. The study asks
whether this representation can improve sequential allocation while retaining a transparent final selector,
and how the answer changes when the reward penalizes DLTs.

This paper makes three contributions. First, it formulates variable-dose Phase I allocation as a
trial-conditioned representation-learning problem with a fixed action space. Second, it evaluates the
resulting policy under a prespecified final MTD-selection rule and matched benchmark designs. Third, it
characterizes the resulting safety--accuracy trade-off across dose counts, reward penalties, learning rates,
and random seeds. Taken together, these contributions provide an empirical framework for studying flexible
allocation under explicit safety criteria, rather than a claim that the current policy dominates conventional
dose-finding designs.

\section{Methods}

Each simulated trial is treated as one episode. First, the simulator samples a monotone dose--toxicity curve
and trial-level design parameters. Patients are then treated in cohorts, and the resulting DLT outcomes
update the observed counts at the current dose. At each decision point, the Transformer converts the
dose-level observations and trial context into a fixed-dimensional representation; the policy uses this
representation to choose de-escalation, staying, or escalation subject to boundary masks. The episode ends
either at an early safety boundary or when the maximum sample size is reached. Only in the latter case does a
prespecified weighted isotonic-regression rule select the final MTD\@. This workflow separates the learned
sequential allocation problem from the final statistical recommendation and provides the structure for the
technical components below.

\subsection{Sequential dose-allocation framework}

Consider ordered dose levels $d_1 < \cdots < d_J$, target DLT probability $\phi$, and maximum sample size
$N_{\max}$. At decision time $t$, dose $j$ is represented by
\begin{equation}
z_{j,t} =
\left(\bar d_j,n_{j,t},x_{j,t},\mathbb{I}\{j=j_t\}\right),
\qquad
\bar d_j = \frac{j-1}{J-1},
\end{equation}
where $n_{j,t}$ and $x_{j,t}$ are the observed patient and DLT counts. The trial query is
\begin{equation}
c_t = \left(\phi,N_{\max},n_t,\bar d_{j_t},J,b\right),
\end{equation}
with $b$ the planned cohort size. The policy observes these quantities but not the simulated truth, including
the underlying toxicity curve or true MTD.

The dose tokens pass through a shared Transformer encoder, following the self-attention architecture
introduced for sequence representation learning \citep{vaswani2017attention}. A query derived from $c_t$
attends over the encoded dose tokens, yielding a fixed-dimensional representation independent of $J$. The use
of shared token processing and aggregation is also related to variable-cardinality representations such as
Deep Sets \citep{zaheer2017deep}; unlike a permutation-invariant set representation, the present model
retains dose order through the normalized dose-position feature. A policy head outputs three actions:
\begin{equation}
\mathcal{A}=\{-1,0,+1\}
=\{\text{de-escalate},\text{stay},\text{escalate}\}.
\end{equation}
At the highest dose, escalation is masked. At the lowest dose, de-escalation remains available and means
early safety stopping with no dose recommended. This boundary action is operationally useful but is not
itself a statistical proof that every dose is unsafe.

Each episode begins at $j_0=1$. After a cohort is treated, the simulator generates individual outcomes
\begin{equation}
Y_{t,r}\sim\operatorname{Bernoulli}(p_{j_t}),\qquad r=1,\ldots,b_t,
\end{equation}
where $b_t=\min\{b,N_{\max}-n_t\}$ allows the final cohort to be smaller than the planned cohort. If
$X_t=\sum_rY_{t,r}$ denotes the cohort DLT count, the observed state is updated by
\begin{equation}
n_{j_t,t+1}=n_{j_t,t}+b_t,\qquad
x_{j_t,t+1}=x_{j_t,t}+X_t.
\end{equation}
The policy is queried only after the cohort outcomes are available. Thus, the simulator may retain the true
curve for evaluation and reward calculation, but the policy receives no information about $p_j$, the true
MTD, future outcomes, or the scenario label.

The action mask is applied before sampling from the policy distribution. De-escalation at the lowest dose is
deliberately not masked because it is the model's explicit early-stop mechanism; escalation at the highest
dose is masked because it would leave the dose grid. These rules yield at most $2J-1$ policy decisions when
$N_{\max}=6J$ and $b=3$, because the last cohort is followed directly by final selection.

\begin{figure}[ht]
\centering
\includegraphics[width=\textwidth]{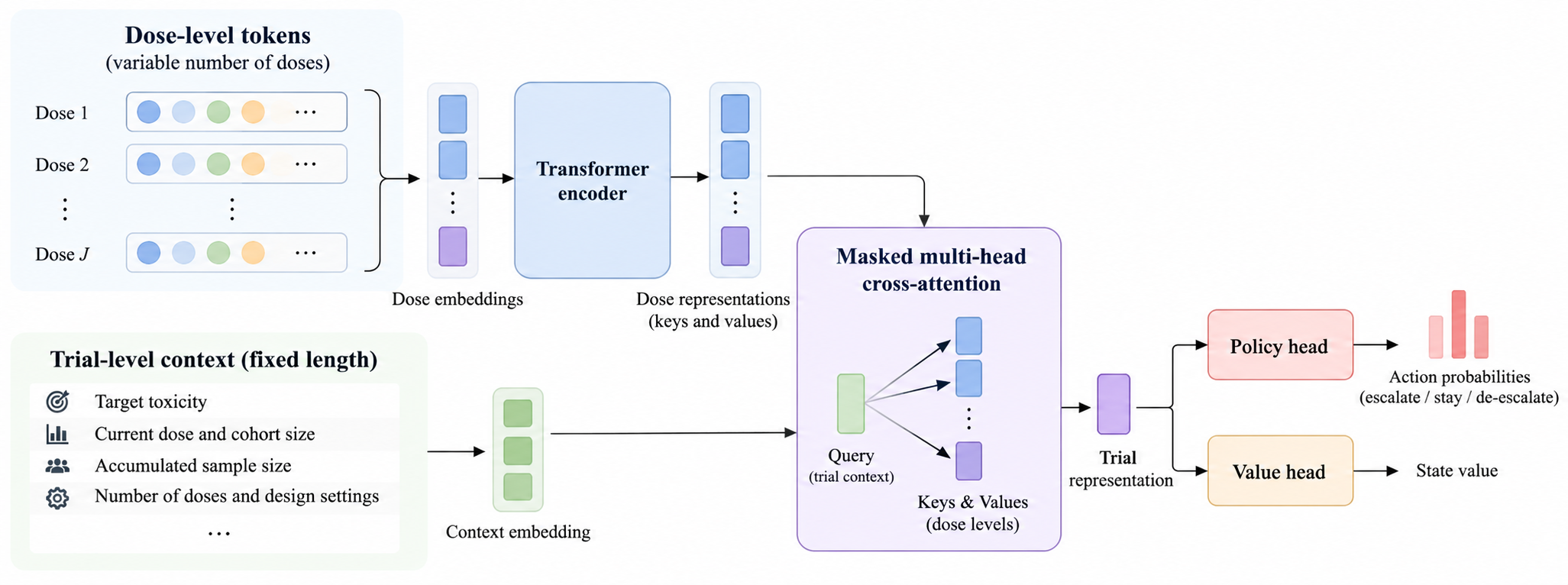}
\caption{Overall Transformer dose-allocation schematic. Variable-length dose-level tokens are encoded into
dose representations, while fixed-length trial context forms the query for masked multi-head cross-attention.
The resulting trial representation feeds the policy and value heads.}
\label{fig:overall-schematic}
\end{figure}

\subsection{Trial-conditioned cross-attention}

As summarized in Figure~\ref{fig:overall-schematic}, the cross-attention layer converts a variable-length
collection of dose representations into one trial representation. Attention was originally introduced as a
learned soft alignment between a query state and relevant elements of an encoded sequence
\citep{bahdanau2015neural}; later work formalized global and local attention variants
\citep{luong2015effective}. In the proposed framework, the dose sequence supplies the information to be
retrieved, while the trial query specifies what information is relevant for the current decision.

Let $H_t\in\mathbb{R}^{J\times p}$ denote the encoded dose tokens and let $q_t\in\mathbb{R}^{p}$ denote the
embedded trial context. Cross-attention produces a fixed-dimensional trial representation
\begin{equation}
h_{\mathrm{trial},t}
=\operatorname{MHA}\!\left(q_t,H_t,H_t;\,M_t\right),
\end{equation}
where the trial context supplies the query and the dose representations supply the keys and values. The mask
$M_t$ excludes padding positions, so the representation is valid for any number of dose levels. The
implementation uses four heads and model dimension $p=64$; the policy and value heads therefore receive the
same-sized input for every $J$.

The purpose of this layer is selective retrieval. For illustration, the architecture can represent a decision
that emphasizes the current dose and nearby toxicity observations early in a trial, or one that emphasizes
more extensively studied doses and evidence near the target later in a trial. These examples describe the
types of state-dependent weighting that cross-attention can express; they are not claims about the attention
patterns necessarily learned by the fitted policy. Unlike masked mean pooling, which gives every observed
dose the same aggregation weight, cross-attention can adapt its weights to the current state. The weights are
learned for action prediction and are not posterior probabilities that a dose is the MTD\@. Standard
query/key/value projections, scaling, and multi-head concatenation are given in
Appendix~\ref{app:attention-details}; the main text focuses on why the mechanism is useful for variable-dose
trials.

\subsection{Prespecified final MTD selection}

If a trial reaches $N_{\max}$, the implementation computes empirical toxicity rates for tried doses and fits
weighted isotonic regression, enforcing the expected monotonicity of toxicity with dose
\citep{barlow1972isotonic,ivanova2009comparison}:
\begin{equation}
(\widetilde p_j)_{j\in\mathcal{O}}
=\arg\min_{q_j\ \mathrm{nondecreasing}}
\sum_{j\in\mathcal{O}}n_j(q_j-\widehat p_j)^2.
\end{equation}

Only doses with $\widetilde p_j\leq\phi+0.05$ are eligible. Among eligible tried doses, the recommendation
minimizes $|\widetilde p_j-\phi|$. Ties within an isotonic block are resolved toward the highest dose when
the fitted value is at or below $\phi$, and toward the lowest dose when it is above $\phi$. If no dose is
eligible, the outcome is no recommendation.

An analysis of the terminal selection procedure indicated that the approximately $0.10$ improvement in PCS
was primarily attributable to the revised tie-breaking rule rather than the safety cap. Therefore, primary
comparisons use the common v2 selection rule where stated.

\subsection{Simulation and training protocol}

The simulation uses monotone dose--toxicity curves with $J\in\{4,6,8,10\}$, target $\phi\in\{0.25,0.30\}$,
cohort size $b=3$, and $N_{\max}=6J$. The primary curves are generated from
\begin{equation}
p_j=\operatorname{logit}^{-1}\left(\alpha+\beta\frac{j-1}{J-1}\right),\qquad \beta>0,
\end{equation}
with parameters varied to produce low-toxicity, target-centered, high-toxicity, plateau, and steep scenarios.
The simulation design also includes monotone nonparametric curves generated from positive random increments
as a stress test of reliance on one logistic functional form. A dose is acceptable when
$p_j\leq\phi_{\mathrm{safe}}$; the primary setting uses $\phi_{\mathrm{safe}}=\phi$. The true MTD is the
acceptable dose closest to $\phi$, with the lowest index used to resolve ties. For the 10\% no-dose
scenarios, the DLT probability at the lowest dose satisfies $p_1>\phi_{\mathrm{safe}}+\delta$ with
$\delta=0.05$, which guarantees that all doses are unsafe while avoiding ambiguous cases immediately adjacent
to the safety boundary.

Each DLT contributes an immediate penalty $-\lambda_{\mathrm{DLT}}$. The terminal outcome receives $+1$ for
correct MTD selection or correct no-dose classification and $-1$ for an incorrect recommendation. Early
stopping at the lowest dose receives $+1$ in a no-dose scenario and $-1$ when an acceptable dose exists.
Consequently, the episode reward can be written as
\begin{equation}
R=R_{\mathrm{terminal}}-\lambda_{\mathrm{DLT}}D_T,
\end{equation}
where $D_T$ is the total number of observed DLTs. For an early-terminated episode, $R_{\mathrm{terminal}}$ is
replaced by $R_{\mathrm{early\,stop}}$. The primary penalty is $\lambda_{\mathrm{DLT}}=0.05$;
$\lambda_{\mathrm{DLT}}=0$ is evaluated as a sensitivity analysis. This decomposition is important because
the scalar reward is an optimization target, whereas PCS, no-dose accuracy, false stopping, and DLT burden
are the clinical operating characteristics.

The main PPO configuration uses a two-layer, four-head Transformer with model dimension 64, four PPO epochs,
generalized advantage estimation parameter $0.95$, discount factor $\gamma=1$, entropy coefficient $0.01$,
and batches of 256 simulated trials. PPO is used because it provides a practical policy-gradient optimization
procedure for the episodic allocation problem \citep{schulman2017ppo}. Training consists of 300 stage-one
iterations followed by 3000 stage-two iterations in the long runs. The value network is trained jointly with
the policy; the final isotonic selector is not learned. Padding masks are kept fixed across batches
containing different values of $J$, and the same action mask is used when computing both action probabilities
and log probabilities.

The simulation separates training, validation, and testing by random seed. Validation scenarios are used for
optimization monitoring and training decisions, while the final evaluation uses 5000 independent
same-distribution test scenarios and greedy policies. The primary analysis reports means and standard
deviations across five RL seeds; optimization-sweep analyses use three seeds. Benchmark designs include BOIN,
i3+3, mTPI-2, and CRM \citep{liu2015boin,liu2020i3,guo2017mtpi2,oquigley1990crm}. Because each benchmark
design has its own toxicity-elimination and final-selection rules, the primary comparison reports both their
original operating characteristics and, where stated, matched evaluations using the common isotonic selector.
This distinction prevents final-selection differences from being attributed incorrectly to the sequential
allocation policy.

The primary test distribution is not intended to establish extrapolation to unseen dose counts. A planned
external-validity analysis would freeze the trained policy and evaluate it on dose counts such as $J=5,7,9$,
shifted target probabilities, altered sample sizes, delayed outcomes, and non-logistic curves. These settings
are therefore treated as generalization targets rather than pooled with the current primary estimates.

\subsection{Operating characteristics and comparator interpretation}

The primary endpoint is the probability of correct selection (PCS), where a no-dose outcome is correct when
no acceptable dose exists. We additionally record the probability that the true MTD was ever tried, PCS
conditional on the true MTD being tried, correct no-dose probability, erroneous early stopping, mean DLT
count, DLT proportion, the fraction of patients assigned above $\phi_{\mathrm{safe}}$, and the distribution
of patients across dose levels. For scenarios with a true MTD, we report POS, the probability that the
selected dose exceeds the true MTD; PCA, the proportion of patients allocated at the true MTD; and POA, the
proportion of patients allocated above the true MTD\@. These three allocation and selection metrics are
undefined in no-dose scenarios and are therefore averaged only over scenarios with an acceptable dose. The
conditional PCS diagnostic is necessary because a final selector cannot recommend an untried true MTD\@.
Unconditional PCS therefore combines an allocation problem with a final-estimation problem.

The benchmark designs are simulated under the same dose grids, cohort size, starting dose, sample-size
budget, and scenario draws. Their original implementations include toxicity-elimination rules, whereas the
primary RL policy uses only its learned boundary stop. In addition, CRM can move more than one dose level in
a single update, while the RL policy and the interval-based benchmark designs move at most one level. These
differences are retained in the primary benchmark because they reflect the complete operating procedures of
the methods. Matched final-selection analyses are used as secondary diagnostics to isolate the effect of the
sequential allocation component.

\section{Results}

\subsection{Safety--accuracy trade-off under toxicity penalties}

We first compared the toxicity-penalized RL policy with the benchmark designs under a common v2
final-selection rule. This experiment tests the primary clinical trade-off: whether reducing DLT exposure is
accompanied by an unacceptable loss of MTD-selection accuracy.

\begin{table}[ht]
\centering
\small
\caption{Primary five-seed v2-selection comparison with a DLT penalty. POS is the probability of selecting
above the true MTD; PCA and POA are the proportions of patients allocated at and above the true MTD,
respectively. RL values are mean $\pm$ standard deviation across seeds; benchmark-design values are single
matched simulation summaries. All methods use the same final selection rule.}
\label{tab:results}
\resizebox{\textwidth}{!}{%
\begin{tabular}{lcccccccc}
\toprule
Method & PCS & POS & PCA & POA & False stop & Mean DLT & Unsafe fraction & Total reward\\
\midrule
RL, $\lambda_{\mathrm{DLT}}=0.05$, lr $10^{-4}$ & $0.561\pm0.017$ & $0.172\pm0.025$ & $0.323\pm0.026$ &
$0.160\pm0.023$ & $0.035\pm0.020$ & $6.51\pm0.57$ & $0.177\pm0.027$ & $-0.130\pm0.016$\\
RL, $\lambda_{\mathrm{DLT}}=0.05$, lr $3\times10^{-5}$ & $0.568\pm0.017$ & $0.155\pm0.020$ & $0.323\pm0.036$
& $0.160\pm0.038$ & $0.021\pm0.013$ & $6.97\pm1.22$ & $0.191\pm0.055$ & $-0.140\pm0.051$\\
BOIN & $0.588$ & $0.220$ & $0.381$ & $0.256$ & $0.008$ & $8.23$ & $0.280$ & $-0.172$\\
i3+3 & $0.589$ & $0.229$ & $0.386$ & $0.272$ & $0.007$ & $8.42$ & $0.293$ & $-0.175$\\
mTPI-2 & $0.591$ & $0.209$ & $0.383$ & $0.248$ & $0.008$ & $8.15$ & $0.271$ & $-0.162$\\
CRM & $0.596$ & $0.211$ & $0.387$ & $0.267$ & $0.008$ & $8.39$ & $0.285$ & $-0.161$\\
\bottomrule
\end{tabular}
}%
\end{table}

The toxicity-penalized policies have lower POS and POA than the benchmark designs, indicating fewer
selections and allocations above the true MTD, while PCA is also lower. Thus, the safety benefit is expressed
not only through aggregate DLT burden but also through more conservative allocation relative to the true
target. The total reward is highest for the penalized RL policies in this comparison because the reward
explicitly values reduced toxicity; it should not be interpreted as a universal clinical performance metric.

\begin{figure}[ht]
\centering
\includegraphics[width=\textwidth]{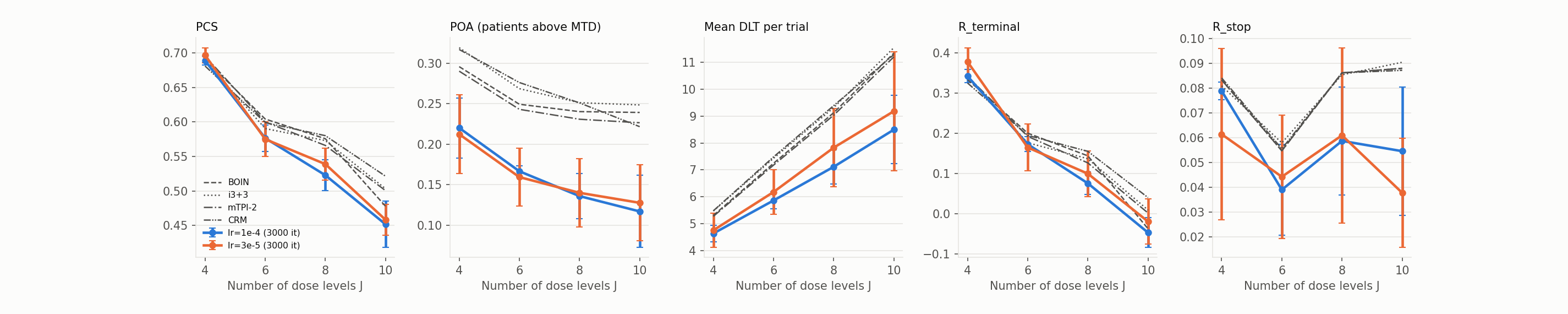}
\caption{Operating characteristics by number of dose levels for the toxicity-penalized RL policies
($\lambda_{\mathrm{DLT}}=0.05$) and matched benchmark designs. The RL curves show means and seed variability;
the benchmark lines are single simulation summaries.}
\label{fig:by-j-lam005}
\end{figure}

The dose-stratified curves indicate that the safety benefit is not uniform across trial sizes: as $J$
increases, the penalized policy becomes increasingly conservative relative to the comparators. This pattern
suggests that dose-grid size remains an important determinant of policy behavior.

\subsection{Behavioral effect of the DLT penalty}

We next removed the DLT penalty while holding the principal simulation and evaluation protocol fixed. This
isolates whether the safety--accuracy pattern is caused by the architecture or by the objective used to train
it. Removing the DLT penalty improves selection accuracy but increases toxicity. The two no-penalty settings
reach PCS values of $0.595$ and $0.605$, comparable to or slightly above the benchmark values, while mean
DLTs increase to approximately 9.1--9.8 and the unsafe-patient fraction increases to approximately
0.32--0.36. False-stop rates are lower without the penalty, at $0.004$--$0.008$.

\begin{table}[ht]
\centering
\small
\caption{Sensitivity of RL operating characteristics to the DLT penalty. POS is the probability of selecting
above the true MTD; PCA and POA are the proportions of patients allocated at and above the true MTD,
respectively. Values are mean $\pm$ standard deviation across five seeds.}
\label{tab:penalty-sensitivity}
\resizebox{\textwidth}{!}{%
\begin{tabular}{lccccccc}
\toprule
Setting & PCS & POS & PCA & POA & False stop & Mean DLT & Unsafe fraction\\
\midrule
$\lambda_{\mathrm{DLT}}=0.05$, lr $10^{-4}$ & $0.561\pm0.017$ & $0.172\pm0.025$ & $0.323\pm0.026$ &
$0.160\pm0.023$ & $0.035\pm0.020$ & $6.51\pm0.57$ & $0.177\pm0.027$\\
$\lambda_{\mathrm{DLT}}=0.05$, lr $3\times10^{-5}$ & $0.568\pm0.017$ & $0.155\pm0.020$ & $0.323\pm0.036$ &
$0.160\pm0.038$ & $0.021\pm0.013$ & $6.97\pm1.22$ & $0.191\pm0.055$\\
$\lambda_{\mathrm{DLT}}=0$, lr $10^{-4}$ & $0.595\pm0.003$ & $0.229\pm0.018$ & $0.352\pm0.032$ &
$0.324\pm0.071$ & $0.004\pm0.003$ & $9.77\pm1.17$ & $0.358\pm0.069$\\
$\lambda_{\mathrm{DLT}}=0$, lr $3\times10^{-5}$ & $0.605\pm0.010$ & $0.186\pm0.022$ & $0.377\pm0.030$ &
$0.287\pm0.054$ & $0.008\pm0.004$ & $9.06\pm0.74$ & $0.316\pm0.057$\\
\bottomrule
\end{tabular}
}%
\end{table}

\begin{figure}[ht]
\centering
\includegraphics[width=\textwidth]{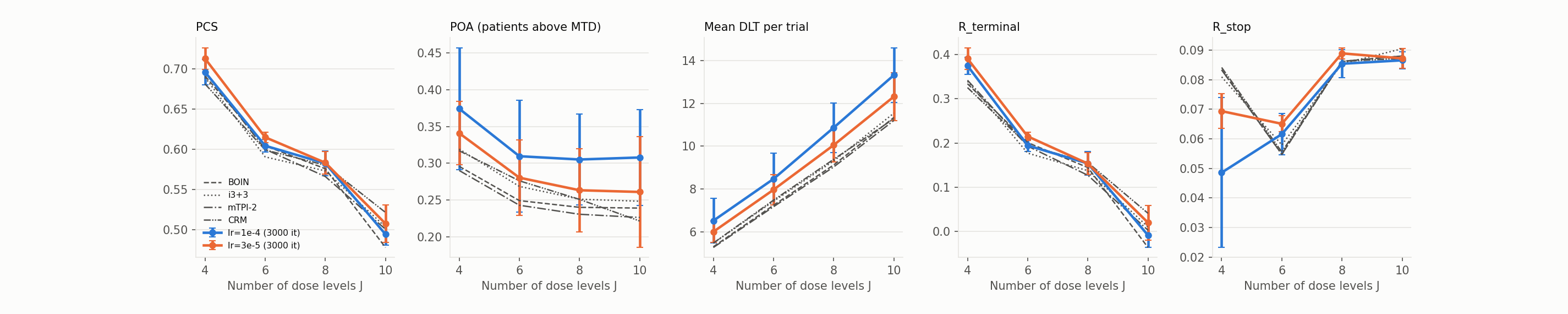}
\caption{Operating characteristics by number of dose levels when the RL DLT penalty is removed
($\lambda_{\mathrm{DLT}}=0$). Selection accuracy increases, but mean DLTs and the fraction of patients
treated above the MTD also increase.}
\label{fig:by-j-lam0}
\end{figure}

The no-penalty curves indicate that improved selection accuracy is achieved through more aggressive
exploration, particularly in larger dose grids. Because the final selection rule is fixed, this pattern is
attributable to allocation behavior rather than to the selector.

\subsection{Performance across dose-grid sizes}

To test the variable-length representation, we examined operating characteristics separately for
$J=4,6,8,10$. Figures~\ref{fig:by-j-lam005} and~\ref{fig:by-j-lam0} show that the policy remains operational
across all four dose counts, but performance is not invariant to $J$. Under the DLT penalty, PCS declines as
the number of dose levels increases, and the RL policies remain below the rule-based reference lines at $J=8$
and $J=10$. Without the penalty, this increase in difficulty as $J$ grows is accompanied by higher DLT burden
and a higher probability of treating above the MTD\@. The variable-length architecture therefore enables
deployment across dose counts, but does not by itself eliminate dose-count-dependent performance differences.

\begin{figure}[ht]
\centering
\includegraphics[width=\textwidth]{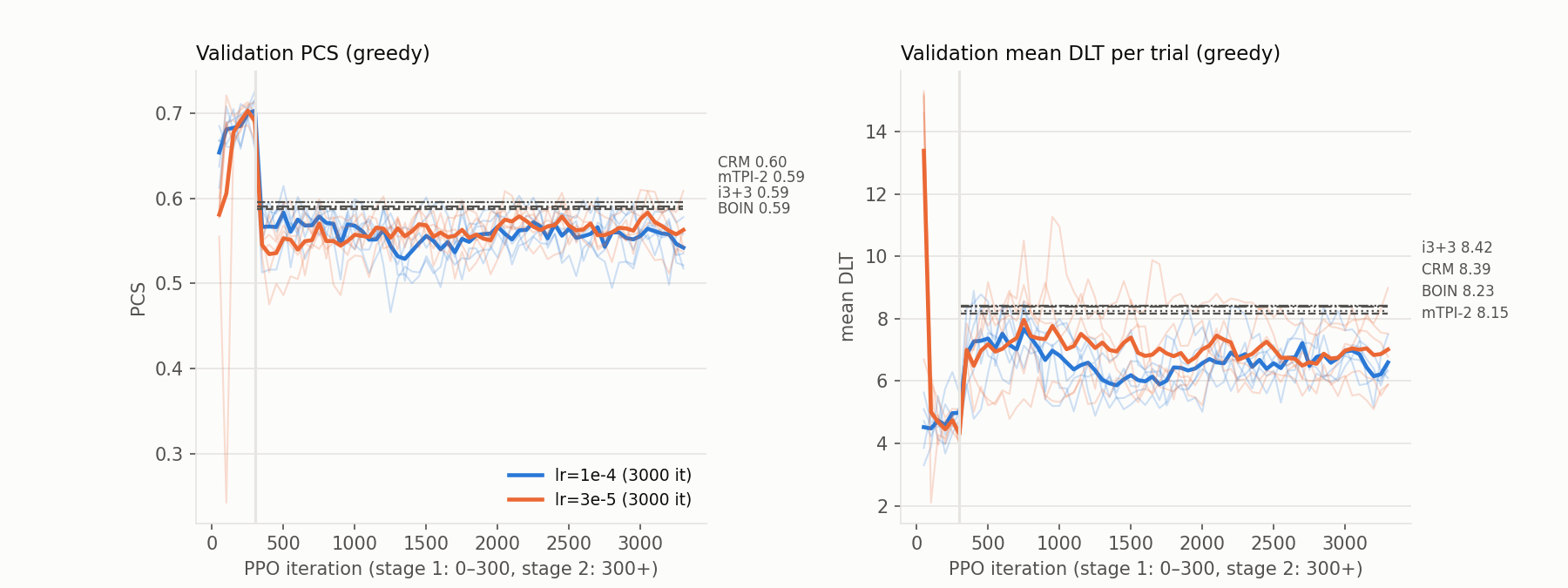}
\caption{Validation training curves for the five-seed long runs with $\lambda_{\mathrm{DLT}}=0.05$. Thin
lines show individual seeds and thicker lines show the aggregate trajectory; horizontal reference lines are
benchmark-design summaries.}
\label{fig:training-curves}
\end{figure}

The training curves show substantial seed-dependent trajectories and no common late-stage improvement,
supporting interpretation of the final policy as a stochastic optimization outcome.

\subsection{Training stability across learning rates and seeds}

We then assessed whether the observed operating characteristics were stable across optimization trajectories.
Figure~\ref{fig:training-curves} shows the validation PCS and mean DLT trajectories for the two five-seed
learning-rate settings with $\lambda_{\mathrm{DLT}}=0.05$. The logged plateau analysis found no significant
late-stage trend in training return or validation PCS for the five-seed long runs. The $10^{-4}$ setting
generally plateaued after roughly 500--1000 stage-two iterations, while $3\times10^{-5}$ plateaued later,
around 1500--2000 iterations. Seed variability is material: for the toxicity-penalized $10^{-4}$ setting, the
observed PCS range across seeds was approximately 0.538--0.581, larger than the estimated test-set sampling
error. The learning-rate comparison therefore does not establish a clear winner; it primarily demonstrates
that seed variation is large enough to require multi-seed reporting.

\subsection{Robustness to PPO optimization settings}

Finally, we examined the completed three-seed sweep over batch size and entropy coefficient. This analysis
asks whether the variability observed above can be reduced through commonly tuned PPO settings. The sweep is
interpreted within the RL conditions; its benchmark rows used their original final-selection rules and are
therefore not pooled with the primary shared-selection comparison.

\begin{table}[ht]
\centering
\small
\caption{Optimization sensitivity sweep. Values are mean $\pm$ standard deviation across three seeds. The
sweep used the original rule-based selection settings for its comparator simulations; only RL rows are shown.}
\label{tab:optimization-sweep}
\resizebox{\textwidth}{!}{%
\begin{tabular}{lccccc}
\toprule
Setting & PCS & False stop & Mean DLT & Unsafe fraction & Total reward\\
\midrule
Baseline: batch 256, entropy 0.01, $\lambda=0.05$ & $0.563\pm0.022$ & $0.026\pm0.022$ & $6.63\pm0.77$ &
$0.181\pm0.035$ & $-0.132\pm0.018$\\
Batch 1024, $\lambda=0.05$ & $0.551\pm0.013$ & $0.044\pm0.006$ & $5.80\pm0.31$ & $0.149\pm0.020$ &
$-0.110\pm0.007$\\
Entropy 0.003, $\lambda=0.05$ & $0.562\pm0.021$ & $0.041\pm0.026$ & $6.36\pm0.81$ & $0.169\pm0.050$ &
$-0.116\pm0.005$\\
Entropy 0.03, $\lambda=0.05$ & $0.552\pm0.007$ & $0.030\pm0.014$ & $6.33\pm0.39$ & $0.167\pm0.031$ &
$-0.133\pm0.014$\\
Baseline: batch 256, entropy 0.01, $\lambda=0$ & $0.595\pm0.004$ & $0.003\pm0.003$ & $9.54\pm0.27$ &
$0.343\pm0.015$ & $-0.226\pm0.019$\\
Batch 1024, $\lambda=0$ & $0.597\pm0.006$ & $0.006\pm0.005$ & $10.04\pm0.79$ & $0.350\pm0.024$ &
$-0.245\pm0.036$\\
Entropy 0.003, $\lambda=0$ & $0.585\pm0.020$ & $0.010\pm0.005$ & $10.03\pm1.72$ & $0.370\pm0.103$ &
$-0.265\pm0.121$\\
Entropy 0.03, $\lambda=0$ & $0.586\pm0.016$ & $0.006\pm0.002$ & $9.58\pm0.30$ & $0.342\pm0.038$ &
$-0.243\pm0.041$\\
\bottomrule
\end{tabular}
}%
\end{table}

\begin{figure}[ht]
\centering
\includegraphics[width=\textwidth]{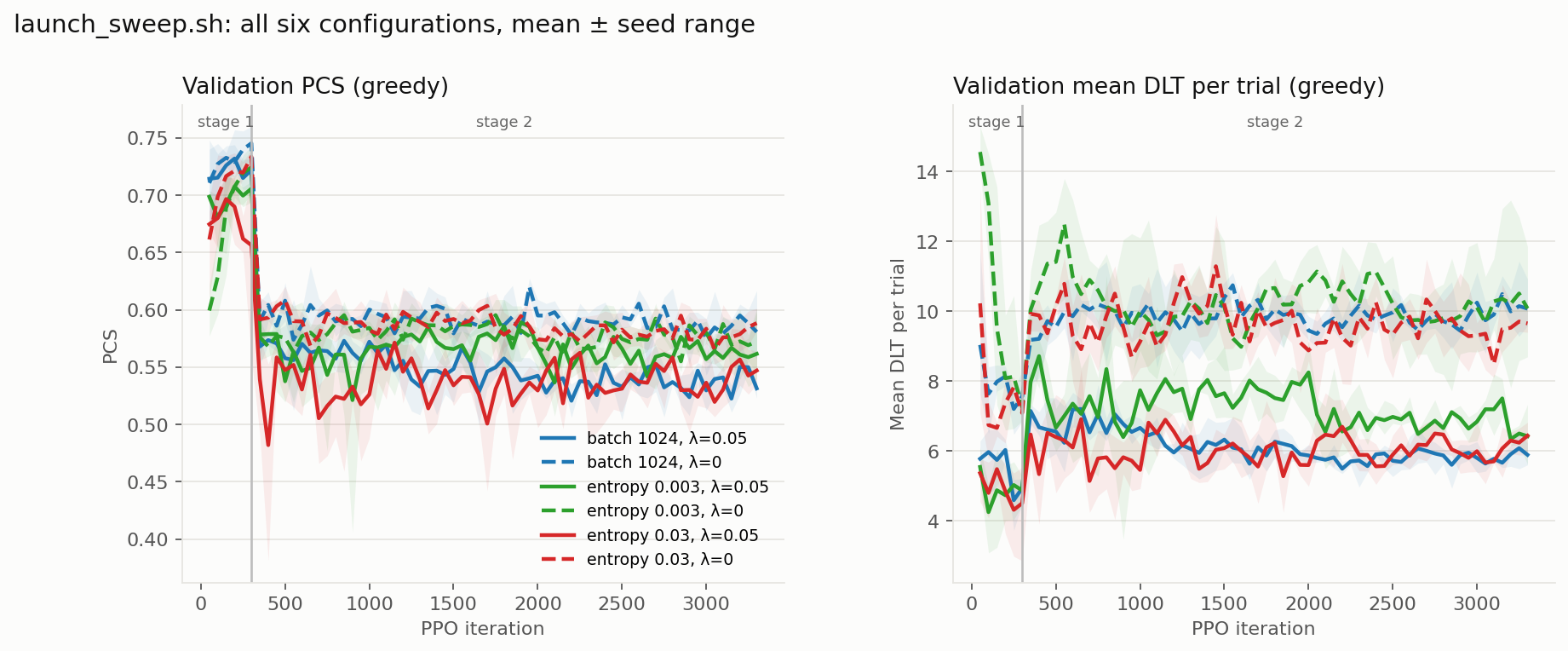}
\caption{Validation PCS and mean DLT trajectories for the completed batch-size and entropy-coefficient sweep.
Shaded regions show seed variability.}
\label{fig:optimization-sweep}
\end{figure}

The sweep does not identify a universally best optimizer setting. The trajectories and final summaries
together suggest that batch size and entropy affect the balance between convergence stability, safety, and
selection accuracy, so optimization sensitivity should be reported as a robustness analysis rather than
reduced to a single reward-maximizing configuration. With a DLT penalty, batch 1024 lowers DLT burden but
also lowers PCS and increases false stopping; without the penalty, it slightly improves PCS while increasing
DLT exposure.

\subsection{Separating allocation and final-selection error}

Overall PCS combines at least two processes: whether the sequential policy ever tries the true MTD, and
whether the final selector chooses it after it has been tried. The aggregate evaluation reports PCS,
false-stop rate, DLT burden, and unsafe-patient fraction, but does not yet report PCS conditional on trying
the true MTD\@. This distinction is important because an untried true MTD is necessarily unavailable to the
isotonic final selector. The present results therefore establish a combined operating characteristic, not a
complete attribution of selection error. A full diagnostic analysis should report the probability that the
true MTD was tried, conditional PCS given that event, correct no-dose probability, and erroneous early-stop
probability.

\subsection{Summary of empirical findings}

Taken together, the experiments show that the policy's operating characteristics depend materially on the
reward penalty, dose-grid size, and optimization seed. The results motivate the more detailed interpretation
and limitations discussed below rather than supporting a claim of uniform superiority over the benchmark
designs.

\section{Discussion}

\subsection{Interpretation of findings}

The principal finding is a controllable trade-off rather than uniform superiority. Penalizing DLTs led the
policy to expose fewer patients to toxicity, but reduced PCS relative to the benchmark designs. Removing that
penalty moved PCS into the benchmark range while increasing DLT burden and unsafe-patient allocation. This
pattern is consistent with the reward definition: the policy is optimizing a different objective when
toxicity carries an immediate cost.

The results also show why the final selector should remain explicit. The isotonic rule and its tie-breaking
convention materially affect PCS, and the current aggregate PCS combines sequential exploration with
final-dose estimation. Keeping this stage prespecified makes the source of the trade-off more inspectable,
but it also means that improvements in the learned allocation policy cannot be inferred from PCS alone.

\subsection{Practical implications}

The framework is best viewed as a simulation platform for evaluating safety-constrained allocation policies.
In a practical development program, PCS, correct no-dose decisions, false-stop probability, mean DLTs, and
the fraction of patients assigned above the MTD should be reported jointly. A scalar episode reward is useful
for optimization but is not sufficient for clinical interpretation.

The variable-length representation is potentially useful when a development program evaluates related dose
grids or when the number of doses differs across protocols. The dose-count stratification nevertheless shows
that parameter sharing does not guarantee equal performance across $J$. Practical deployment would also
require prespecified operating-characteristic thresholds and trial-design software and workflow support
\citep{zhou2021boinsuite}. The toxicity penalty is a soft safety constraint rather than a formal guarantee;
safe reinforcement-learning methods and constrained policy-optimization approaches provide relevant
methodological directions \citep{garcia2015comprehensive,achiam2017constrained}.

\subsection{Limitations}

The study is simulation-based and currently evaluates same-distribution test scenarios. Logistic monotone
curves may not capture all clinically relevant dose--toxicity shapes, and pending outcomes are not modeled.
Time-to-event extensions such as TITE-BOIN explicitly address late-onset toxicity and incomplete follow-up
\citep{yuan2018titeboin,chen2025titeboin}; adapting the present RL environment to those settings is an
important limitation and future extension. The RL policy does not yet use an independent overdose-protection
rule or a statistical toxicity-elimination rule, whereas the benchmark designs include their own safety
mechanisms. The reported benchmark-design values are single simulation runs, whereas RL values average over
seeds; the comparison is therefore informative but not perfectly symmetric. The completed optimization sweep
uses three seeds and a different comparator-selection convention from the primary analysis.

\subsection{Future work}

Future simulation studies should evaluate non-logistic and shifted scenario families, unseen dose counts,
multiple independent replicates for every benchmark design, and the allocation/final-selection decomposition
of PCS\@. Probability-of-decision designs offer another strategy for accelerating Phase I decisions and could
inform future extensions of the allocation environment \citep{zhou2020pod}. A clinically richer extension
could add pharmacokinetic and pharmacometric information to the trial context and dose tokens, building on
recent pharmacometrics-enabled dose-optimization work for seamless Phase I--II oncology trials
\citep{yuan2026pharmacometrics}. Another extension could allow backfill enrollment, so that patients can be
assigned to previously explored lower doses while escalation continues, as in the backfill i3+3 design
\citep{liu2023backfill}. These additions would require explicit modeling of information availability, cohort
timing, and the safety rules governing concurrent dose assignments. Additional ablations should isolate the
Transformer encoder, cross-attention, and variable-dose training. Finally, the DLT penalty should be
calibrated against an explicitly stated safety or utility criterion rather than selected solely by total
episode reward.

In conclusion, this study presents a variable-length Transformer reinforcement-learning framework for
sequential Phase I dose allocation with trial-conditioned cross-attention and a transparent, prespecified
final MTD selector. The simulations show that toxicity penalties reduce DLT exposure but trade away some
selection accuracy, whereas removing the penalty produces more accurate but more aggressive allocation. The
framework provides a structured platform for evaluating safety-constrained policies against benchmark designs
and for future extensions incorporating richer clinical information.

\appendix
\section{Standard cross-attention details}
\label{app:attention-details}

For completeness, the multi-head cross-attention used in the policy is specified here. For head $m$, the
embedded trial query and encoded dose tokens are projected as
\begin{equation}
Q_t^{(m)}=q_tW_Q^{(m)},\qquad
K_t^{(m)}=H_tW_K^{(m)},\qquad
V_t^{(m)}=H_tW_V^{(m)}.
\end{equation}
The attention weights are
\begin{equation}
\alpha_t^{(m)}
=\operatorname{softmax}\left(
\frac{Q_t^{(m)}K_t^{(m)\top}}{\sqrt{d_k}}+M_t
\right),
\end{equation}
where $M_t$ assigns a large negative value to padded dose positions. The head output is
$u_t^{(m)}=\alpha_t^{(m)}V_t^{(m)}$, and the multi-head output is
\begin{equation}
h_{\mathrm{trial},t}
=\operatorname{Concat}\left(
u_t^{(1)},\ldots,u_t^{(M)}
\right)W_O.
\end{equation}
The implementation uses four heads, model dimension $p=64$, and a shared encoder across dose counts. The
no-cross-attention ablation replaces this operation with masked mean pooling of the encoded dose tokens
concatenated with the query embedding.

\bibliographystyle{apalike}
\IfFileExists{references.bib}{\bibliography{references}}{\bibliography{manuscript/references}}

\end{document}